\documentclass[conference]{IEEEtran}
\IEEEoverridecommandlockouts

\usepackage{cite}
\usepackage{amsmath,amssymb,amsfonts}
\usepackage{graphicx}
\usepackage{xcolor}
\usepackage{graphicx} %
\graphicspath{{figures/}} %

\usepackage{amsmath,amsfonts,bm}

\def\eqref#1{equation~\ref{#1}}

\def\1{\bm{1}}

\def\vx{{\bm{x}}}

\DeclareMathAlphabet{\mathsfit}{\encodingdefault}{\sfdefault}{m}{sl}
\SetMathAlphabet{\mathsfit}{bold}{\encodingdefault}{\sfdefault}{bx}{n}

\newcommand{\vt}{\bm{\tau}}
\newcommand{\vth}{\bm{\theta}}

\usepackage[hidelinks]{hyperref}
\usepackage{cleveref}

\crefname{figure}{Fig.}{Figs.}
\Crefname{figure}{Figure}{Figures}
\crefname{equation}{Eq.}{Eqs.}
\Crefname{figure}{Equation}{Equations}
\crefname{section}{Sec.}{Secs.}
\Crefname{section}{Section}{Sections}
\crefname{table}{Tab.}{Tabs.}
\Crefname{table}{Table}{Tables}

\usepackage{booktabs}
\usepackage{multirow}
\usepackage{tabularx}

\usepackage{colortbl} %
\newcolumntype{Y}{>{\centering\arraybackslash}X}
\usepackage{nicematrix}
\usepackage{fontawesome5}

\usepackage{graphicx}
\usepackage{pgfplots}
\pgfplotsset{compat=1.18}
\usetikzlibrary{calc}

\usepackage{enumitem}

\usepackage{xcolor}
\usepackage{amsmath, amssymb}

\definecolor{colorA}{HTML}{0f8b8d}
\definecolor{colorB}{HTML}{fa9426}
\definecolor{colorTable}{HTML}{F2E3DF}
\definecolor{TVmerge}{HTML}{412F39}
\definecolor{TVa}{HTML}{D4786C}
\definecolor{TVb}{HTML}{915D71}
\usepackage{tikz}
\usetikzlibrary{arrows.meta, calc, positioning}
\def\BibTeX{{\rm B\kern-.05em{\sc i\kern-.025em b}\kern-.08em
    T\kern-.1667em\lower.7ex\hbox{E}\kern-.125emX}}

\usetikzlibrary{fit, shapes.symbols, positioning}

\definecolor{ink}{RGB}{58,47,54}
\definecolor{sourcefill}{RGB}{241,226,223} %
\definecolor{vectorfill}{RGB}{231,194,186} %
\definecolor{mergefill}{RGB}{215,128,116}  %

\begin{document}

\title{Robust Zero-Shot Generalization for Open-Vocabulary Action Recognition\\ via Task Arithmetic}

\author{
\IEEEauthorblockN{
Francesca Morandi$^{\star\,1,2}$, Omayma Moussadek$^{\star\,1}$, Federico Venturini$^3$, Mauro Suardi$^3$,\\[0.2cm]
Alessandro Banzatti$^3$, Francesco Cannarile$^3$, Angelo Porrello$^1$, Simone Calderara$^1$
}\\

\IEEEauthorblockA{
\begin{tabular}{ccc}
$^1$University of Modena and Reggio Emilia &
$^2$University of Pisa &
$^3$Eni S.p.A. \\
\texttt{$^1$\{name.surname\}@unimore.it} &
\texttt{$^2$\{name.surname\}@phd.unipi.it} &
\texttt{$^3$\{name.surname\}@eni.com}
\end{tabular}\\[0.15cm]
$^\star$Equal contribution.
}
}

\maketitle

\begin{abstract}
Open Vocabulary Action Recognition (OVAR) enables the recognition of novel actions by leveraging vision–language representations, overcoming the limitations of traditional closed-set approaches. However, achieving robust performance in real-world scenarios typically requires domain-specific fine-tuning, which is often costly and raises privacy and regulatory concerns. In this work, we propose an alternative paradigm that bypasses target-domain training and recombines knowledge from existing datasets and models. Leveraging model merging and task arithmetic, we extract and combine task vectors from models fine-tuned on diverse public OVAR datasets. We show that, in out-of-distribution settings, the resulting merged model achieves superior zero-shot generalization to the pre-trained base model. Code is available at \url{https://github.com/omaymaMoussadek/robust-ovar}
\end{abstract}

\section{Introduction}
\label{sec:intro}
Action recognition is a fundamental task in computer vision that aims to identify human actions from video data, with applications in surveillance, human-computer interaction, autonomous driving, and video retrieval. Traditionally, it is formulated as a \textit{closed-set classification problem}, where models are trained on a fixed set of predefined action classes and cannot generalize to unseen ones. In contrast, \textbf{Open Vocabulary Action Recognition (OVAR)} leverages multimodal vision-language representations~\cite{radford2021learning} to align videos with textual descriptions, enabling the recognition of novel actions at inference time by matching visual inputs with semantic concepts expressed in language. This multimodal approach effectively shifts the paradigm from a closed-set to an \textit{open-dictionary setting}, where actions can be specified through structured prompts that encode richer information than simple predicates, while also allowing the recognition of actions not observed during training, thus providing greater flexibility.

Although the \textbf{zero-shot} capabilities of OVAR models theoretically allow them to operate in new domains without further training, achieving robust performance in real-world scenarios often requires a \textbf{fine-tuning phase} on domain-specific data. This is particularly evident in complex and heterogeneous environments characterized by poor lighting conditions, multiple interacting actors and objects, and distant or low-resolution camera views. In these contexts, a reliable recognition system requires data-driven adaptation.

This requirement poses a significant challenge, as collecting and annotating video data for action recognition and video analytics is inherently \textbf{sensitive} and \textbf{difficult}. In many applications, data acquisition involves continuous monitoring of individuals in physical spaces, raising \textbf{serious privacy concerns}. Moreover, such systems fall under strict regulatory frameworks: for instance, the AI Act classifies many of these applications as \textbf{high-risk}, due to their reliance on biometric data, video surveillance, and the potential for behavioral profiling. As a result, acquiring in-domain data for fine-tuning is often not only costly, but also legally and ethically constrained.

To overcome the intractability of standard fine-tuning for OVAR, we explore an alternative direction: instead of adapting models to the target domain, we recombine and fuse the knowledge already embedded in existing, safely trained models. This strategy aligns with a broader trend in artificial intelligence based on the reuse and combination of existing models, that has become particularly relevant in fields such as image classification and natural language processing~\cite{wortsman2022model, matena2022merging}, where the increasing scale and complexity of deep architectures makes fine-tuning progressively impractical.

\begin{figure}[t]
    \centering
\begin{tikzpicture}[scale=0.5, >=Stealth]

\coordinate (T0A) at (1.1,1.1);
\coordinate (a1)  at (2.1,1.4);
\coordinate (a2)  at (3.2,2.4);
\coordinate (a3)  at (4.2,1.5);
\coordinate (TA)  at (5.4,2.6);

\draw[dashed, very thick, black]
    (T0A)
    .. controls (1.6,1.0) and (2.0,1.2) .. (a1)
    .. controls (2.6,2.0) and (2.9,2.6) .. (a2)
    .. controls (3.8,2.4) and (4.0,1.3) .. (a3)
    .. controls (4.8,1.2) and (5.1,2.1) .. (TA);

\draw[-{Stealth[open, length=4mm, width=3mm]}, line width=2pt, TVa]
    (T0A) -- (TA)
    node[midway, below, sloped] {$\vt_A$};

\draw[fill=gray!20, draw=black, line width=1.0pt] (T0A) circle (5pt);
\draw[fill=gray!20, draw=black, line width=1.0pt] (TA) circle (5pt);

\node[below left=-1pt] at (T0A) {$\vth_0$};
\node[above right=0pt] at (TA) {$\vth_A$};

\node at (3.5,0.2) {
    {\normalsize $f(\vx, \vth_0 + \textcolor{TVa}{\vt_A})$}
};

\begin{scope}[xshift=1.5cm]
\coordinate (T0B) at (6.8,1.1);
\coordinate (TB)  at (11.6,1.2);

\draw[dashed, very thick, black]
    (T0B)
    .. controls (7.4,0.7) and (7.8,0.6) .. (8.4,1.1)
    .. controls (8.8,1.6) and (9.2,1.7) .. (9.8,1.2)
    .. controls (10.2,0.7) and (10.6,0.6) .. (TB);

\draw[-{Stealth[open, length=4mm, width=3mm]}, line width=2pt, TVb]
    (T0B) -- (TB)
    node[midway, below] {$\vt_B$};

\draw[fill=gray!20, draw=black, line width=1.0pt] (T0B) circle (5pt);
\draw[fill=gray!20, draw=black, line width=1.0pt] (TB) circle (5pt);

\node[below left=2pt] at (T0B) {$\vth_0$};
\node[above right=2pt] at (TB) {$\vth_B$};

\node at (9.2,2.5) {
    {\normalsize $f(\vx, \vth_0 + \textcolor{TVb}{\vt_B})$}
};
\end{scope}

\begin{scope}[shift={($(T0A)!0.5!(TB) + (-3.0cm,-6cm)$)}]

\coordinate (T0Sum) at (0,1.1);

\coordinate (vA_rel) at (2.625, 1.975);
\coordinate (vB_rel) at (3.0, 0.075);

\coordinate (TASum) at ($(T0Sum) + (vA_rel)$);
\coordinate (TBSum) at ($(T0Sum) + (vB_rel)$);
\coordinate (TSum)  at ($(T0Sum) + (vA_rel) + (vB_rel)$);

\draw[dashed, thick, gray!60] (TASum) -- (TSum);
\draw[dashed, thick, gray!60] (TBSum) -- (TSum);

\draw[-{Stealth[open, length=4mm, width=3mm]}, line width=2pt, TVa]
    (T0Sum) -- (TASum)
    node[midway, above left] {$\vt_A$};

\draw[-{Stealth[open, length=4mm, width=3mm]}, line width=2pt, TVb]
    (T0Sum) -- (TBSum)
    node[midway, below] {$\vt_B$};

\draw[-{Stealth[length=4mm, width=4mm]}, line width=3pt, TVmerge]
    (T0Sum) -- (TSum);

\draw[fill=gray!20, draw=black, line width=1.0pt] (T0Sum) circle (5pt);
\draw[fill=gray!20, draw=black, line width=1.0pt] (TASum) circle (5pt);
\draw[fill=gray!20, draw=black, line width=1.0pt] (TBSum) circle (5pt);
\draw[fill=gray!20, draw=black, line width=1.0pt] (TSum) circle (5pt);

\node[below left] at (T0Sum) {$\vth_0$};
\node[above right=3pt] at (TASum) {$\vth_A$};
\node[below right] at (TBSum) {$\vth_B$};
\node[right=3pt] at (TSum) {$\vth_{\text{TA}}$};

\node at (3.2,-0.4) {
    {\normalsize $f(\vx, \vth_0 + \textcolor{TVa}{\vt_A} + \textcolor{TVb}{\vt_B})$}
};

\end{scope}

\end{tikzpicture}
\caption{Given two models fine-tuned from the same weights, task addition~\cite{ilharco2022editing} produces a multi-task model by summing their corresponding task vectors.}
    \label{fig:task_arithmetic}
\end{figure}
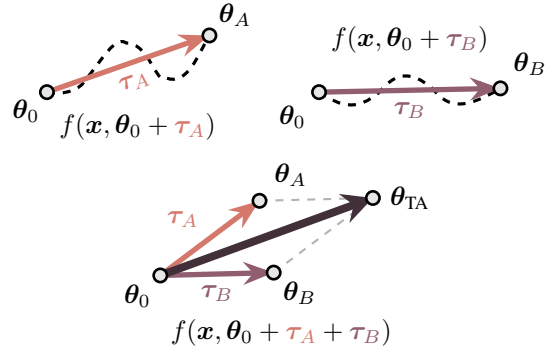

Central to this line of research is task arithmetic~\cite{ilharco2022editing, ortiz2023task}, which enables model editing and merging through simple algebraic operations in parameter space. This approach relies on the concept of \textbf{task vector}, formally defined as the exact displacement in parameter space between a pre-trained base model and its task-specific fine-tuned counterpart. As demonstrated in recent literature~\cite{yadav2023ties, wang2024localizing, gargiulo2025task}, these vectors isolate learned knowledge so it can be directly manipulated: subtracting them enables the selective removal of behaviors, while summing them allows models sharing the same initialization to be merged into a single multi-task network (as shown in~\cref{fig:task_arithmetic}).

The main objective of our work is to deploy a modular and scalable approach for open-vocabulary action recognition that operates entirely without target-domain fine-tuning. By leveraging task arithmetic and advanced model merging techniques, we achieve robust zero-shot generalization, maintaining high performance even in out-of-distribution (OOD) settings characterized by severe distribution shifts with respect to the pre-trained model. To achieve this, we leverage a collection of fine-tuned models trained on diverse, publicly available datasets~\cite{carreira2019short, soomro2012ucf101, 6126543, wu2020not}, from which we extract the corresponding task vectors. These task vectors are then combined through addition, resulting in a unified multi-task model that aggregates knowledge from multiple sources. In this way, the model acquires more refined and diverse action understanding capabilities, improving generalization across a broader range of action classes without further in-domain training.

While the effectiveness of model merging techniques has been widely demonstrated in Large Language Models (LLMs)~\cite{yu2024language, yadav2023ties}, image classification~\cite{ilharco2022editing}, and medical imaging~\cite{lumetti2025u}, their evaluation in the video domain remains  unexplored. This leaves a significant gap in understanding the impact of subspace merging and task arithmetic on the complex spatiotemporal representations required for open-vocabulary action recognition. We show that, even in the absence of a specialized target dataset, it is possible to successfully merge models fine-tuned on diverse, publicly available OVAR datasets. Crucially, the resulting models outperform the original zero-shot baseline, even when the source datasets are entirely out-of-distribution with respect to the target application, mitigating the need for sensitive, domain-specific data collection. To sum up, our main contributions are:
\begin{enumerate}
    \item \textbf{Novel application to video:} We present the first systematic evaluation of existing model merging and task arithmetic techniques (including Task Arithmetic (TA)~\cite{ilharco2022editing}, TSV-M~\cite{gargiulo2025task}, and Iso-C~\cite{marczak2025no}) applied specifically to Open Vocabulary Action Recognition.
    \item \textbf{Privacy-preserving OOD adaptation:} We provide empirical evidence that parameter-space merging of models fine-tuned on public datasets reliably yields constructive interference and generalized performance gains in the video domain, even under out-of-distribution shifts. By achieving effective target-domain adaptation entirely through task vector aggregation, our approach completely eliminates the need to collect or train on sensitive, application-specific video data.
\end{enumerate}

\section{Background}
\label{sec:background}
\subsection{Open-Vocabulary Video Recognition}
Contrastive vision-language models like CLIP\cite{radford2021learning} exhibit strong zero-shot classification ability by aligning visual and textual representations within a shared embedding space. This allows the model to recognize categories beyond the training set by matching visual features with text descriptions. 

However, extending this paradigm from images to videos is not straightforward, as action recognition requires modeling temporal dynamics in addition to visual semantics. \textbf{Open-VCLIP} \cite{weng2023open} addresses this limitation by adapting CLIP to video recognition while preserving its open-vocabulary nature. A key aspect of Open-VCLIP is the introduction of lightweight temporal modeling directly inside the visual transformer. Instead of processing each frame independently, the model expands the self-attention operation so that visual patches can attend not only to patches from the same frame, but also to those in adjacent frames, including both preceding and subsequent ones. In this way, CLIP’s original spatial attention is transformed into a spatio-temporal mechanism able to capture short-range temporal dependencies without adding extra parameters.

\subsection{Model Soups and Task Arithmetic}
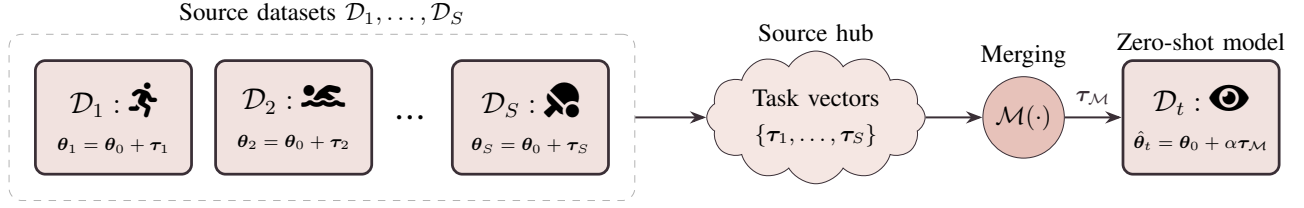
\begin{figure*}[t]
    \centering
    \resizebox{0.95\textwidth}{!}{
        \begin{tikzpicture}[
            node distance=0.6cm,
            box/.style={
                draw=ink,
                rectangle,
                rounded corners,
                minimum width=2.2cm, 
                minimum height=1.6cm,
                align=center,
                line width=1.1pt,
                font=\small
            },
            source_box/.style={
                draw=gray!60,
                dashed,
                inner sep=10pt,
                rounded corners=5pt
            },
            hub_style/.style={
                cloud,
                cloud puffs=12,
                cloud puff arc=120,
                aspect=2.0,
                draw=ink,
                fill=vectorfill!40,
                minimum width=2.5cm,
                minimum height=1.5cm,
                align=center
            },
            arrow/.style={-Stealth, thick, draw=ink},
            merge_node/.style={
                circle,
                draw=ink,
                fill=vectorfill,
                inner sep=3pt,
                font=\normalsize
            },
            lbl/.style={
                font=\small\itshape,
                text=ink,
                align=center,
                inner sep=1pt
            }
        ]

        \node[box, fill=sourcefill] (D1) {
            {\large $\mathcal{D}_1 : $} {\Large \faRunning} \\[5pt]
            \scriptsize $\vth_1 = \vth_0 + \vt_1$
        };
        
        \node[box, fill=sourcefill, right=0.3cm of D1] (D2) {
            {\large $\mathcal{D}_2 : $} {\Large \faSwimmer} \\[5pt]
            \scriptsize $\vth_2 = \vth_0 + \vt_2$
        };

        \node[right=0.2cm of D2] (dots) {\textbf{\dots}};

        \node[box, fill=sourcefill, right=0.2cm of dots] (DS) {
            {\large $\mathcal{D}_S : $} {\Large \faTableTennis} \\[5pt]
            \scriptsize $\vth_S = \vth_0 + \vt_S$
        };

        \node[source_box, fit=(D1) (D2) (dots) (DS), label=above: {Source datasets $\mathcal{D}_1,\dots,\mathcal{D}_S$}] (container) {};

        \node[hub_style, right=1cm of container, label=above:Source hub] (hub) {Task vectors\\[2pt] \small \{$\vt_1, \dots, \vt_S$\}};
        
        \draw[arrow] (container.east) -- (hub.west);

        \node[merge_node, right=0.8cm of hub, label=above:Merging] (merge) {$\mathcal{M}(\cdot)$};
        \draw[arrow] (hub.east) -- (merge);

        \node[box, fill=sourcefill, right=0.8cm of merge, label=above:Zero-shot model] (target) {
            {\large $\mathcal{D}_t : $} {\Large \faEye} \\[5pt]
            \scriptsize $\hat{\vth}_t = \vth_0 + \alpha \vt_{\mathcal{M}}$
        };
        \draw[arrow] (merge) -- (target)
            node[midway, above=4pt, lbl] {$\vt_{\mathcal{M}}$};

        \end{tikzpicture}
    }
    \caption{Experimental protocol for zero-shot evaluation. Source models are fine-tuned on individual datasets $\mathcal{D}_1, \dots, \mathcal{D}_S$. Task vectors are aggregated in the \textit{Source Hub}, merged, and then evaluated on the held-out target $D_t$.}
    \label{fig:pipeline}
\end{figure*}

In recent years, model soups~\cite{wortsman2022model} have shown that averaging the weights of fine-tuned models originating from the same pre-trained initialization can improve both accuracy and robustness without increasing inference cost. This approach was originally studied in a setting where multiple models are fine-tuned on the same downstream task and dataset, differing only in training hyperparameters such as learning rate, data order, or regularization. Empirically, weight averaging in this setting yields a model that matches in-domain performance while significantly improving robustness under distribution shifts. 

Task arithmetic~\cite{ilharco2022editing} represents updates as the parameter difference between a fine-tuned model and its pre-trained initialization. These parameter differences, called task vectors, can be manipulated through simple operations such as addition, subtraction, and scaling, enabling controlled modifications of model behavior without retraining from scratch.

A key difference with respect to model soups is that, in task arithmetic, each model is fine-tuned on a different dataset or objective, and the resulting task vectors encode distinct capabilities. Prior work~\cite{ortiz2023task,yoshida2025mastering,porrello2026dataless} has shown that, under a linearized regime, \textit{i.e.}, when models are approximated via a first-order Taylor expansion around the pre-trained parameters, task vectors exhibit structured compositionality. Complementarily, recent second-order analyses of model compositionality suggest that composable task-specific modules can also emerge in standard non-linear networks, provided that fine-tuning remains within the pre-training basin~\cite{porrello2025second}. Related work~\cite{sommariva2026distilling} bridges linearized and standard non-linear fine-tuning by distilling the representations of a linearized teacher into a non-linear student.

In this paper, we build on these insights and employ task arithmetic to fuse knowledge from multiple datasets by combining their task vectors. Compared to centralized retraining, this approach offers a more scalable and modular alternative, enabling incremental integration of new capabilities without access to all training data or full model retraining.
\subsection{Advanced Merging Strategies}
\label{sec:adv}
While task vector addition offers a simple and efficient way to combine models, its effectiveness can be limited when the source models are specialized for different domains or tasks. In such cases, interactions between parameter updates may attenuate useful task-specific directions. Prior work shows that this can hinder the preservation of performance achieved by individually trained models, as conflicting updates may lead to degradation across tasks~\cite{gargiulo2025task,marczak2025no}. To overcome this issue, more advanced model merging strategies leverage the internal structure of weight updates instead of merging the network as a single flat vector. Among them, \textbf{Task Singular Vector (TSV-M)}~\cite{gargiulo2025task} and \textbf{Iso-C}~\cite{marczak2025no} tackle the problem from two complementary perspectives:
\begin{itemize}
  \item \textbf{TSV-M}~\cite{gargiulo2025task} performs a structured merge based on singular value decomposition. For each matrix-shaped parameter, it analyzes task updates at the layer level, extracts the most informative low-rank directions, and recombines them into a merged parameter update. Non-matrix parameters are merged by direct averaging.
  \item \textbf{Iso-C}~\cite{marczak2025no} starts from the average task vector and then applies an isotropization step, replacing the original singular value spectrum with a flatter one. This reduces dominant directions that may bias the merged updates toward a subset of source tasks, leading to a more balanced combination of shared and task-specific information.
\end{itemize}

Driven by these findings, we analyze whether advanced merging strategies, such as TSV-M and Iso-C, can improve performance in out-of-domain settings, \textit{i.e.}, when transferring to datasets that differ from those used during fine-tuning.

\section{Task Arithmetic for Robust Open-VOC Action Recognition}
\label{sec:method}

Given a publicly available dataset $\mathcal{D} = \{(V_i, T_i)\}_{i=1}^N$, where each video clip is paired with a textual description, we initialize the model from a pre-trained Open-VCLIP checkpoint, denoted by $f_{\vth_0}$ with parameters $\vth_0$~\cite{weng2023open}, and fine-tune it to learn a shared embedding space in which each video is aligned with its corresponding text. Let $f_{\vth^{(V)}}(\cdot)$ and $f_{\vth^{(T)}}(\cdot)$ denote the visual and text encoders, respectively. For a video-text pair $(V,T)$, the corresponding embeddings are
\begin{align}
v = f_{\vth^{(V)}}(V), \qquad t = f_{\vth^{(T)}}(T),
\end{align}
and their alignment is encouraged by maximizing the inner-product similarity $\mathrm{sim}(v,t) = \langle v, t \rangle$. During fine-tuning, the text encoder is kept frozen~\cite{ilharco2022editing}; therefore, the optimization primarily adapts the visual encoder. At inference time, class predictions are obtained by matching visual video representations against text-derived class prototypes built from a fixed set of CLIP-style prompt templates. 

After fine-tuning on a source dataset $\mathcal{D}$, we extract the \textbf{task vector} of the visual encoder as
\begin{equation}
\vt = \vth^{(V)} - \vth_0^{(V)},\quad \text{where }\vth^{(V)} = \mathcal{A}(\vth_0^{(V)}, \mathcal{D}),
\end{equation}

where $\vth_0$ denotes the pre-trained model (\textit{initialization}) and $\mathcal{A}(\cdot)$ is a gradient-based adaptation procedure.
\subsection{Source Model Adaptation}
A common limitation of standard fine-tuning is the lack of access to target-domain training data, often due to privacy constraints or data-sharing restrictions. A natural alternative is to leverage existing datasets from related domains. In principle, multiple publicly available action recognition datasets could be combined into a single large training set for joint fine-tuning.

However, this approach is often impractical in real-world scenarios, since datasets typically differ in data formats, annotation protocols, class vocabularies, video lengths, frame sampling strategies, and dataset-specific preprocessing pipelines (\textit{e.g.}, data loaders, augmentations, and normalization schemes). Harmonizing these aspects into a unified training framework requires engineering effort and may still result in suboptimal performance due to domain inconsistencies.

To address these challenges, we adopt a \textbf{modular strategy}. We assume access to a collection of \emph{source} action recognition datasets $\{\mathcal{D}_s\}_{s=1}^{S}$, while the \emph{target} dataset $\mathcal{D}_t$ is not available for training. Instead of merging all datasets at the data level, we independently fine-tune the same pre-trained model $\vth_0$ on each source dataset $\mathcal{D}_s$, obtaining a set of source-specific parameters $\{\vth_s\}_{s=1}^{S}$. Each of these models captures domain-specific adaptations, which we represent through the corresponding hub of task vectors $\mathcal{T}_{\texttt{Source-Hub}}$:
\begin{equation}
\mathcal{T}_{\texttt{Source-Hub}} = \{\vt_s \, | \, \vt_s = \vth_s^{(V)} - \vth_0^{(V)} \}_{s=1}^{S} 
\end{equation}
A schematic overview of the proposed method is provided in~\cref{fig:pipeline}, which illustrates the extraction of task vectors from source-specific models, their subsequent merging, and zero-shot evaluation on the held-out target dataset.
The extracted source task vectors are then combined into a single update, as described in the following subsection.

\subsection{Task-Vector Merging}

Given the set of source task vectors $\mathcal{T}_{\texttt{Source-Hub}} = \{\vt_s\}_{s=1}^{S}$, we define a merging function $\mathcal{M}(\cdot)$ that aggregates them into a single update vector $\vt_{\mathcal{M}}$. As a baseline, we consider standard Task Arithmetic (TA), i.e.,
\begin{equation}
    \vt_{\mathcal{M}} = \mathcal{M}\big(\mathcal{T}_{\texttt{Source-Hub}}\big),
    \qquad
    \textit{e.g., } \vt_{\mathcal{M}} = \sum_{s=1}^{S} \vt_s \;\;
\end{equation}
In addition to TA, we assess the advanced merging strategies Iso-C~\cite{marczak2025no} and TSV-M~\cite{gargiulo2025task} introduced in~\cref{sec:adv}. The resulting merged update is used to construct the target model as $\hat{\vth}_t = \vth_0 + \alpha \, \vt_{\mathcal{M}}$ where $\alpha$ is a scaling coefficient controlling the contribution of the merged task vector to the base model.

\section{Experiments}
\label{sec:experiments}

In the following, we evaluate the ability of different merging strategies (TA, Iso-C, TSV-M) to generalize in an out-of-distribution setting, \textit{i.e.}, on a held-out external dataset, while also assessing their capacity to preserve in-domain performance, \textit{i.e.}, on the source datasets used during adaptation.

\subsection{Experimental setting} 
\label{sec:setting}

\begin{table}[t]
    \centering
    \caption{Overview of the datasets used in our experiments.}
    \label{tab:datasets_overview}
    \setlength{\tabcolsep}{3pt}
    \scriptsize
    \newcommand{\tallrow}{\rule[-1ex]{0pt}{3.5ex}} 
    \begin{NiceTabularX}{\columnwidth}{l c c Y}
        \toprule      
        \specialrule{0pt}{0pt}{0pt}
        \rowcolor{colorTable} \tallrow \textbf{Dataset} & \textbf{\#Classes} & \textbf{\#Videos} & \textbf{Description} \\ 
        \specialrule{0pt}{0pt}{0pt}
        \midrule
        K700~\cite{carreira2019short}   & 700 & 650,317 & Daily activities, YouTube clips\\
        UCF101~\cite{soomro2012ucf101}  & 101 & 13320   & Sports and human actions\\
        HMDB51~\cite{6126543}           & 51  & 6766    & Cinematic and facial motions\\
        XD-Violence~\cite{wu2020not}    & 7   & 4754    & Audio-visual violent events\\
        \bottomrule
    \end{NiceTabularX}
\end{table}

\noindent\textbf{Backbones.} For all experiments, we adopt Open-VCLIP~\cite{weng2023open} as the base model and consider two visual backbones, ViT-B/16 and ViT-L/14. Both architectures are initially adapted to the video domain via pre-training on Kinetics-400 (K400)~\cite{kay2017kinetics}. We employ the official public checkpoints provided by the Open-VCLIP authors~\cite{weng2023open} and maintain their standard training and inference protocols across all experiments.

\smallskip
\noindent\textbf{Datasets.} We evaluate our approach on four action recognition benchmarks. Kinetics-700 (K700)~\cite{carreira2019short} is a large-scale collection of curated YouTube clips covering a broad range of daily activities and social interactions. UCF101~\cite{soomro2012ucf101} features realistic videos across five macro-groups, including human and object interactions and sports in unconstrained environments. HMDB51~\cite{6126543} is a cinematic dataset composed of clips extracted from movies and public databases, focusing on facial and body movements. Lastly, XD-Violence~\cite{wu2020not} is a massive multi-modal benchmark for anomaly detection, capturing complex violent events through both RGB and audio signals. \Cref{tab:datasets_overview} summarizes their main characteristics.

\begin{figure}[tb]
  \centering
  \includegraphics[width=\linewidth]{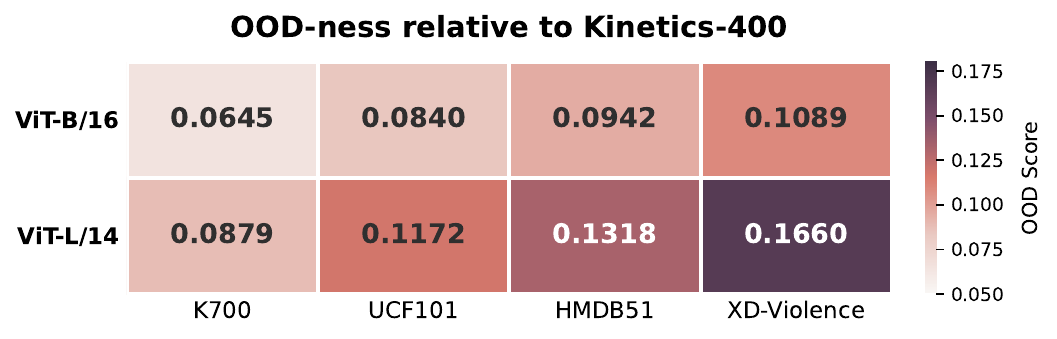}
\caption{For each dataset, we report its out-of-distribution (OOD) shift relative to K400, used to pre-train the base model. Larger values (\textit{e.g.}, for XD-Violence) indicate a stronger semantic mismatch with the source domain.}
\vspace{-0.8em}
\label{fig:heatmap_ood}
\end{figure}

\smallskip
\noindent\textbf{Evaluation Protocol.} To rigorously assess zero-shot generalization, we evaluate all merging methods under a \textbf{leave-one-dataset-out protocol} across HMDB51, UCF101, K700, and XD-Violence. In each of the four iterations, one dataset is held out as the target (test) domain. The merged model is then constructed by extracting and aggregating task vectors from the models fine-tuned on the remaining three source datasets. Crucially, the training split of the target dataset is never observed during this process, ensuring a pure out-of-distribution evaluation.

\smallskip
\noindent\textbf{Implementation Details.} All models are optimized using AdamW~\cite{loshchilov2017decoupled} with a fixed learning rate of $3.33 \times 10^{-6}$. The performance of the merged model is governed by a scaling factor $\alpha$, which modulates the contribution of the task-specific vectors. Following the original setups of~\cite{ortiz2023task, ilharco2022editing}, we determine the optimal $\alpha$ by performing a grid search in the range $[0.1, 2.0]$ with a step size of $0.1$, selecting the value that yields the highest accuracy on the target dataset's validation split.

\smallskip
\noindent\textbf{Metrics.} To assess overall performance, we report Top-1 and Top-5 accuracy of the models on the target dataset.
\begin{figure*}[t]
  \centering
  \includegraphics[width=0.81\linewidth]{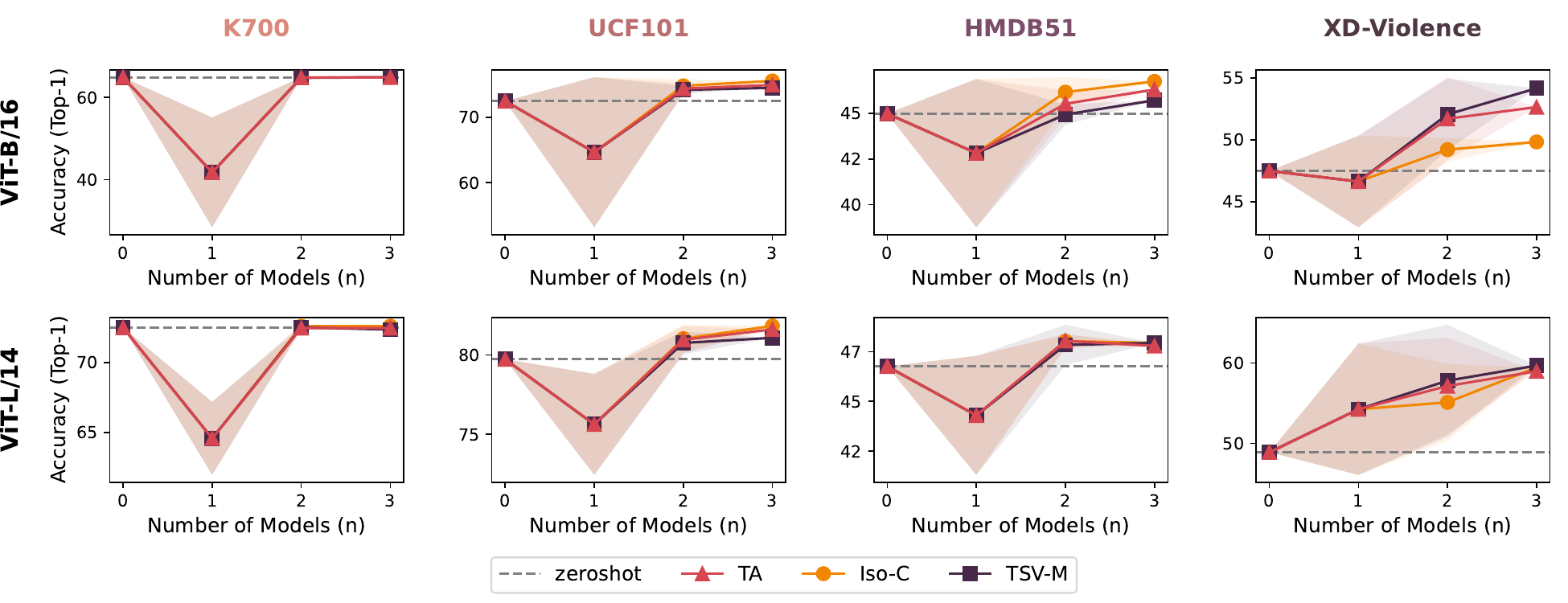}
\caption{Target accuracy versus number of fused source models, showing how performance scales as more task vectors are aggregated.}
\vspace{-0.8em}
\label{fig:increasing}
\end{figure*}

\subsection{Quantifying Distribution Shifts} 
To evaluate the ability of different merging strategies to generalize in an out-of-distribution (OOD) setting, we must first quantitatively assess the semantic gap exhibited by each target dataset with respect to the base model. We achieve this by adopting a text-based metric derived from the class labels of the source domain (\textit{i.e.}, K400) and those of the target domain.

Specifically, we extract the embeddings of the class labels of K400 and each target dataset using the CLIP text encoder, and compute all pairwise cosine similarities between the corresponding textual embeddings. For each target label, we retain the maximum similarity to any K400 label and average these values over the entire set of target labels. We then define the OOD score as one minus this mean maximum similarity, such that higher values indicate a larger semantic mismatch with respect to K400:
\begin{equation}
\mathrm{OOD}(\mathcal{T}) = 1 - \frac{1}{|\mathcal{T}|} \sum_{t \in \mathcal{T}} \max_{c \in \mathcal{C}} \ \mathrm{sim}\big(f_{\vth^{(T)}}(t), f_{\vth^{(T)}}(c)\big)
\end{equation}
where $\mathcal{T}$ denotes the set of class labels in the target dataset, $\mathcal{C}$ the set of class labels in K400 and $\mathrm{sim}\big(f_{\vth^{(T)}}(t), f_{\vth^{(T)}}(c)\big)$ the cosine similarity between the textual embedding of target label $t$ and that of source label $c$.

We report the OOD scores of each target dataset in~\cref{fig:heatmap_ood} for both ViT-B/16 and ViT-L/14. Overall, K700 is the closest dataset to K400, whereas \textbf{XD-Violence is the most distant}. In contrast, HMDB51 and UCF101 remain relatively close to K400, indicating a broader semantic alignment. We leverage these observations to contextualize the following results.
\begin{table}[t]
    \centering
    \caption{Results under the leave-one-dataset-out protocol. For each dataset, merging strategies are evaluated using task-specific checkpoints from the remaining three source domains.}
    
    \label{tab:resultstab1}
    \setlength{\tabcolsep}{2pt} %
    \scriptsize %
    \newcommand{\tallrow}{\rule[-1ex]{0pt}{3.5ex}} 
    \begin{tabularx}{\columnwidth}{ll *{8}{Y}}
        \toprule
        & & \multicolumn{2}{c}{K700} & \multicolumn{2}{c}{UCF101} & \multicolumn{2}{c}{HMDB51} & \multicolumn{2}{c}{XD-Violence} \\
        \cmidrule(lr){3-4} \cmidrule(lr){5-6} \cmidrule(lr){7-8} \cmidrule(lr){9-10}
        Model & Strategy & Top-1 & Top-5 & Top-1 & Top-5 & Top-1 & Top-5 & Top-1 & Top-5 \\
        \midrule
        \specialrule{0pt}{0pt}{0pt}
        \rowcolor{colorTable} \multicolumn{10}{c}{\tallrow\textbf{ViT-B/16}} \\ 
        \specialrule{0pt}{0pt}{0pt}
        \midrule
        \multirow{1}{*}{Fine-tuned} & & 78.38 & 93.76 & 99.47 & 99.92 & 80.97 & 93.81 & 83.35 & 99.18 \\
        \midrule
        \multirow{1}{*}{Zero-shot} & & 64.87 & 86.37 & 72.45 & 89.56 & 44.99 & 69.76 & 47.48 & 91.44 \\
        \midrule
        \multirow{3}{*}{Merged} & TA  & 64.89 & \textbf{86.19} & 74.92 & 89.79 & 46.31 & 69.32 & 52.64 & 94.72 \\
                                & Iso-C & 64.85 & 86.08 & \textbf{75.53} & \textbf{90.09} & \textbf{46.76} & 69.76 & 49.82 & 92.61 \\
                                & TSV-M & \textbf{64.91} & 86.00 & 74.47 & 89.94 & 45.72 & \textbf{70.80} & \textbf{54.16} & \textbf{94.73} \\
        
        \midrule
        \specialrule{0pt}{0pt}{0pt}
        \rowcolor{colorTable} \multicolumn{10}{c}{\tallrow\textbf{ViT-L/14}} \\
        \specialrule{0pt}{0pt}{0pt}
        \midrule
        \multirow{1}{*}{Fine-tuned} & & 82.77 & 95.47 & 99.77 & 100.00 & 83.92 & 95.87 & 84.64 & 99.41 \\
        \midrule
        \multirow{1}{*}{Zero-shot} & & 72.48 & 90.18 & 79.73 & 93.60 & 46.76 & 71.98 & 48.89 & 92.03 \\
        \midrule
        \multirow{3}{*}{Merged} & TA  & 72.45 &  \textbf{90.31} & 81.61 &  93.32 & 47.79 & 72.86  & 58.97 & 94.61 \\
                                  & Iso-C & \textbf{72.59} &  90.26 & \textbf{81.83} &  93.62 & \textbf{47.94} & 73.01  & 59.44 & \textbf{96.48} \\
                             & TSV-M & 72.50 &  90.27 & 81.08 &  \textbf{93.84} & \textbf{47.94} & \textbf{73.60}  & \textbf{59.67} & 95.43 \\
        
        \bottomrule
    \end{tabularx}
\end{table}

\subsection{Zero-Shot Generalization in Open-Voc Action Recognition}
To understand how varying degrees of distribution shift impact actual performance, we evaluate our approach under the leave-one-dataset-out protocol detailed in \cref{sec:setting}. By iteratively holding out each dataset as the unseen target, we analyze how performance scales when merging different combinations of 1, 2, and 3 source models fine-tuned on the remaining domains, reporting average accuracy across ensembles. As illustrated in~\cref{fig:increasing}, increasing the pool of merged models positively correlates with performance on OOD distributions, yielding results consistently superior to those of the original zero-shot model. This suggests that merging multiple models integrates complementary knowledge, a factor that is beneficial for overcoming the base model's limitations. 

Furthermore, performance gains appear to be correlated with the ``OOD-ness'' of the dataset: the more distant the downstream task is from the pre-trained model (\textit{e.g.}, XD-Violence), the greater the benefits of increasing the number of merged models. Conversely, when the task is closely related to the pre-training domain, the improvements are negligible, as the base model already exhibits strong inherent proficiency.

Next, we evaluate the performance obtained when utilizing the maximum number of source models per target dataset, specifically by merging the checkpoints fine-tuned on all the remaining datasets. As presented in~\cref{tab:resultstab1}, more complex fusion strategies, such as Iso-C and TSV-M, provide tangible benefits when the target dataset is distant from the pre-trained model, whereas they perform comparably to simple task vector summation when the knowledge gap is minimal. Furthermore, the efficacy of model merging remains consistent across different backbones (ViT-B/16 and ViT-L/14).
\begin{table}[t]
    \centering
    \caption{Hyperparameter-diverse Pools (XD-Violence): ``Merged'' and ``All-8 Merged'' fuse 1 and 8 models per dataset, respectively.}
\label{tab:results_soup}
    \setlength{\tabcolsep}{4pt}
    \scriptsize
    \newcommand{\tallrow}{\rule[-1ex]{0pt}{3.5ex}} 
    \begin{tabular}{ll cc c cc}
        \toprule
        \specialrule{0pt}{0pt}{0pt}
        & & \multicolumn{2}{c}{\cellcolor{colorTable}\tallrow \textbf{ViT-B/16}} 
        & %
        & \multicolumn{2}{c}{\cellcolor{colorTable}\tallrow \textbf{ViT-L/14}} \\
        
        \specialrule{0pt}{0pt}{0pt}
        \cmidrule(l{0.0\tabcolsep}r{0.0\tabcolsep}){3-4}
        \cmidrule(l{0.0\tabcolsep}r{0.0\tabcolsep}){6-7}
        Model & Strategy & Top-1 & Top-5 & & Top-1 & Top-5 \\
        \midrule

        Fine-tuned &   & 83.35 & 99.18 & & 84.64 & 99.41 \\
        Zero-shot   &   & 47.48 & 91.44 & & 48.89 & 92.03 \\
        \midrule

        \multirow{3}{*}{Merged}
            & TA  & 52.64 & 94.72 & & 58.97 & 94.61\\
            & Iso-C & 49.82 & 92.61 & & 59.44 & \textbf{96.48} \\
            & TSV-M & \textbf{54.16} & \textbf{94.73} & & \textbf{59.67} & 95.43 \\
        \midrule

        \multirow{3}{*}{ALL-8 Merged}
            & TA  & 53.81 & 94.49 & & 64.24 & 96.72 \\
            & Iso-C & 49.82 & 93.02 & & 65.53 & \textbf{97.30} \\
            & TSV-M & \textbf{54.75} & \textbf{95.43} & & \textbf{66.59} & 96.13 \\
        \bottomrule
    \end{tabular}
\end{table}

\smallskip
\noindent\textbf{Leveraging Hyperparameter-Diverse Model Pools.} 
Inspired by prior work on model soups~\cite{wortsman2022model}, which shows that averaging checkpoints obtained from different hyperparameter configurations can improve robustness, we extend this idea to a multi-task merging setting to further expand knowledge diversity. Instead of relying on a single fine-tuned model per non-held-out source dataset under the same leave-one-dataset-out protocol, we generate \textbf{eight distinct source checkpoints} per dataset to be used for subsequent merging. These checkpoints are obtained by varying fine-tuning hyperparameters, including learning rate, weight decay, and data augmentation.

The resulting checkpoints are first merged within each dataset via simple averaging, yielding a single consolidated checkpoint per dataset. These consolidated checkpoints are then combined using TA, Iso-C, and TSV-M, as in the previous subsection. We evaluate this two-level approach on XD-Violence, using the validation split to tune the scaling factor $\alpha$ and the test split to report the final results. As shown in~\cref{tab:results_soup}, increasing source diversity at the hyperparameter level produces more transferable merged models under distribution shifts, yielding substantial gains over the single-model merging approach discussed previously. This further corroborates our intuition that leveraging model diversity is essential for composing a model that is robust against distribution shifts.
\subsection{Model Merging Performance}
Finally, we evaluate the performance of the merged models on the union of the source datasets (using their respective test sets). This is consistent with standard evaluation practices in model merging literature~\cite{ilharco2022editing}, where the objective is to assess whether the fused model preserves the knowledge embedded in each fine-tuned model comprising the ensemble.

We report the results in~\cref{tab:resultstab_indomain}. As shown, except for UCF101, the performance of the merged model is generally inferior to that of individual fine-tuning. Nevertheless, the application of more complex merging strategies is generally rewarding, suggesting that there remains scope for improving merging strategies for open-vocabulary action recognition.

\section{Conclusion}
\label{sec:conclusion}
\begin{table}[t]
    \centering
    \caption{Comparison under standard merging evaluation protocol.}
    \label{tab:resultstab_indomain}
    \setlength{\tabcolsep}{2pt} %
    \scriptsize %
    \newcommand{\tallrow}{\rule[-1ex]{0pt}{3.5ex}} 
    \begin{tabularx}{\columnwidth}{ll *{8}{Y}}
        \toprule
        & & \multicolumn{2}{c}{K700} & \multicolumn{2}{c}{UCF101} & \multicolumn{2}{c}{HMDB51} & \multicolumn{2}{c}{XD-Violence} \\
        \cmidrule(lr){3-4} \cmidrule(lr){5-6} \cmidrule(lr){7-8} \cmidrule(lr){9-10}
        Model & Strategy & Top-1 & Top-5 & Top-1 & Top-5 & Top-1 & Top-5 & Top-1 & Top-5 \\
        \midrule
        \specialrule{0pt}{0pt}{0pt}
        \rowcolor{colorTable} \multicolumn{10}{c}{\tallrow \textbf{ViT-B/16}} \\
        \specialrule{0pt}{0pt}{0pt}
        \midrule
        \multirow{1}{*}{Fine-tuned} & & 78.38 & 93.76 & 99.47 & 99.92 & 80.97 & 93.81 & 83.35 & 99.18 \\
        \midrule
        \multirow{1}{*}{Zero-shot} & & 64.87 & 86.37 & 72.45 & 89.56 & 44.99 & 69.76 & 47.48 & 91.44 \\
        \midrule
        \multirow{3}{*}{Merged} & TA  & 67.01 & 88.03 & 99.55 & 100.00 & 71.39 & 91.30 & 78.43 & 98.59 \\
                                & Iso-C & \textbf{68.96} & \textbf{89.87} & \textbf{99.77} & \textbf{100.00} & 72.42 & 91.15 & 75.62 & 99.06 \\
                               & TSV-M & 67.28 & 88.50 & \textbf{99.77} & \textbf{100.00} & \textbf{74.48} & \textbf{91.59} & \textbf{80.19} & \textbf{99.06} \\
        
        \midrule
        \specialrule{0pt}{0pt}{0pt}
        \rowcolor{colorTable} \multicolumn{10}{c}{\tallrow \textbf{ViT-L/14}}\\
        \specialrule{0pt}{0pt}{0pt}
        \midrule
        \multirow{1}{*}{Fine-tuned} & & 82.77 & 95.47 & 99.77 & 100.00 & 83.92 & 95.87 & 84.64 & 99.41 \\
        \midrule
        \multirow{1}{*}{Zero-shot} & & 72.48 & 90.18 & 79.73 & 93.60 & 46.76 & 71.98 & 48.89 & 92.03 \\
        \midrule
        \multirow{3}{*}{Merged} & TA  & 79.45 & 94.60 & 98.05 & 99.85  & 71.09  & 91.30 & \textbf{84.29} & 99.18 \\
                                & Iso-C & \textbf{80.13} & \textbf{95.18} & 99.02 & 99.92  & 73.89  & 92.04 & 83.59 & \textbf{99.41} \\
                               & TSV-M & 80.11 & 94.85 & \textbf{99.77} & \textbf{100.00}  & \textbf{74.50}  & \textbf{93.36} & 84.06 & 99.06 \\
        
        \bottomrule
    \end{tabularx}
\end{table}

We investigate the application of task arithmetic to open-vocabulary action recognition as a means to address performance degradation under distribution shifts. We demonstrate that the efficacy of these merging strategies is positively correlated with the semantic distance (OOD-ness) of the target task, with advanced fusion techniques providing tangible benefits. Furthermore, we increase model diversity by extending this approach to pools of models trained with different hyperparameters, showing that such diversity enhances model transferability and robustness. Collectively, our results suggest that model merging offers a promising paradigm for efficient, adaptable, and generalizable action recognition systems.

\bibliographystyle{IEEEtran}
\bibliography{reference}

\end{document}